%% file: acl_latex.tex
\pdfoutput=1

\documentclass[11pt]{article}

\usepackage{multirow}
\usepackage{graphicx}
\usepackage{booktabs}
\usepackage{colortbl}
\usepackage{paralist}
\usepackage{enumitem}
\usepackage{amssymb}
\usepackage{hyperref}
\usepackage{bm}
\usepackage{makecell}
\usepackage{arydshln}
\usepackage{pifont}
\usepackage{longtable}
\usepackage{acl}
\usepackage{times}
\usepackage{latexsym}
\usepackage[T1]{fontenc}
\usepackage[utf8]{inputenc}
\usepackage{microtype}
\usepackage{inconsolata}

\title{Multimodal Fake News Video Explanation: Dataset, Analysis and Evaluation}

\newcommand{\emailsI}{\href{mailto:20234027010@stu.suda.edu.cn}{20234027010}}

\author{
Lizhi Chen$^{1}$\quad
Zhong Qian$^{1}$\thanks{$^*$Corresponding author.}\quad
Peifeng Li$^{1}$\quad
Qiaoming Zhu$^{1}$ \\
	$^{1}$School of Computer Science and Technology Soochow University \\
	\texttt{\{\emailsI\}@stu.suda.edu.cn}\\
}

\usepackage{multirow}

\begin{document}
\maketitle
\begin{abstract}
Multimodal fake news videos are difficult to interpret because they require comprehensive consideration of the correlation and consistency between multiple modes. Existing methods deal with fake news videos as a classification problem, but it's not clear why news videos are identified as fake. Without proper explanation, the end user may not understand the underlying meaning of the falsehood. Therefore, we propose a new problem - Fake news video Explanation (FNVE) - given a multimodal news post containing a video and title, our goal is to generate natural language explanations to reveal the falsity of the news video. To that end, we developed FakeVE, a new dataset of 2,672 fake news video posts that can definitively explain four real-life fake news video aspects. In order to understand the characteristics of fake news video explanation, we conducted an exploratory analysis of FakeVE from different perspectives. In addition, we propose a Multimodal Relation Graph Transformer (MRGT) based on the architecture of multimodal Transformer to benchmark FakeVE. The empirical results show that the results of the various benchmarks (adopted by FakeVE) are convincing and provide a detailed analysis of the differences in explanation generation of the benchmark models.
\end{abstract}

\section{Introduction}
Fake news detection has become an important social issue \cite{1, 2}, especially in the digital era where information spreads extremely fast. Traditionally, fake news detection mainly focuses on the content of the text and judges the veracity of the news by analyzing the potential false information in the text. However, with the rise of video platforms, the manifestation of fake news has become more complex, and based on the wide application of multimodal content \cite{4, 5} in news communication, its persuasibility and propagation speed are far superior to plain text news. Therefore, it becomes crucial to use multimodal signals for Fake News Video Detection (FNVD). Existing studies \cite{6,7,8} primarily utilize heterogeneous multimodal information in news content for the detection and identification of fake news, by capturing inter- and intra-modal inconsistencies.

\begin{figure}[t]
	\centering
	\includegraphics[width=1\linewidth]{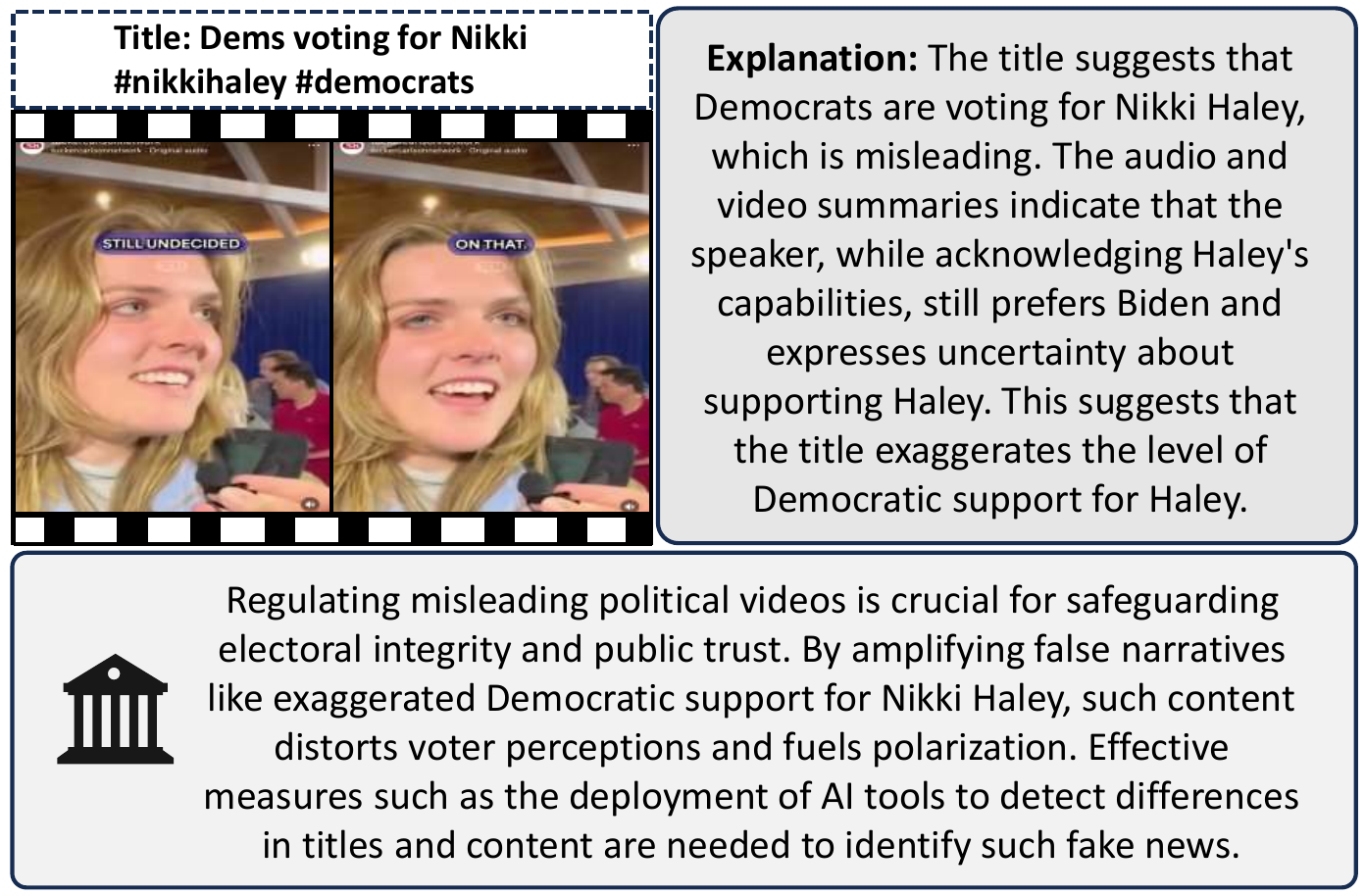}
	\caption{Example scenarios of multimodal fake news video explanation task. Through the explanation to reveal the false reasons behind it, targeted measures are taken to prevent and combat.}
    	\vspace{-0.3cm}
	\label{fig:f1}
\end{figure}

Fake news videos often use the manipulation of video clips \cite{9} or elaborate news title and audio \cite{10} to make the presented content seriously inconsistent with the facts, thus misleading the audience's cognition. In reality, only the veracity of news videos is often difficult to meet the needs of practical applications \cite{11}. Understanding why these news videos are false and revealing the manipulation and motivation behind them through explanation \cite{12} is crucial for effectively detecting and responding to malicious public opinion. This is because only when we truly understand how fake news videos are created can we take targeted measures to prevent and combat them. Figure \ref{fig:f1} shows a false example of video content inconsistent with news titles. From the explanation, we learn that news distorts voters' opinions and exacerbates polarization by exaggerating the Democratic Party's support for Nikki Haley. Therefore, we can target and detect such misleading political videos, which is critical to maintaining election integrity and public trust.

To this end, we propose a new question - Fake News Video Explanation (FNVE). The task takes multiple modes (text, audio, video) as input and aims to generate natural language sentences to interpret falsity in news videos. FNVE tasks differ from traditional explanation systems. Whereas traditional explanation systems often rely on heat maps of attention \cite{13} or other visualization mechanisms \cite{14} to explain model decisions, FNVE aims to generate coherent, fine-grained natural language explanations to accurately describe fake content in news videos. This task paradigm shift makes the explanation of false information more intuitive and understandable, helping to improve the transparency and credibility of fake news detection systems.

Most work examining the explanation of fake news \cite{3,15} has focused on fake news based on text or images, while very little work has explained fake news in the form of video. The existing work has two main limitations: 1) Dataset: The previous work \cite{16,19,24} constructed FNVD dataset, although the veracity label of news videos was marked, but the lack of fine-grained explanation of the specific falsification methods and manipulation motives of fake videos led to the limited application value of FNVD dataset and the difficulty in verifying its rationality. 2) Generation method: Most of the work still relies on the preliminary stage of manual feature extraction, which is difficult to deal with complex scenarios such as news video tampering. Moreover, the potential of pre-training generation models (such as multimodal Transformer) in feature learning is not fully utilized, and automated explanation generation frameworks are lacking. And then hinder the development and practical application of the benchmark model of FNVE task.

To bridge the gap in the FNVE mission, we propose an English dataset containing complete fine-grained fake news video explanations (FakeVE for short). The dataset consists of 2,672 fake news video posts with natural language explanatory sentences manually annotated by expert annotators, Accurate coverage of contextual dishonesty, splice tampering, synthetic voiceover and contrived absurdity, four types of high-incidence fake news video aspects. By exploring and analyzing FakeVE from the aspects of explanation content distribution and explanation error, we demonstrate the complex characteristics of FNVE, and provide key implications for the design of subsequent explanation generation methods.

To solve the problem of automatic generation of FNVE, we propose a Multimodal Relation Graph Transformer (MRGT) based on the architecture of multimodal Transformer. MRGT introduces multi-modal relationship graph to comprehensively represent the relationship of multi-modes (i.e., news title, video frame, audio transcript) to promote the reasoning of fake news videos. Finally, a BARt-based decoder was added to the pipeline to interpret the generation. The experimental results show that MRGT has a significant advantage in capturing the multimodal feature association of video and generating logically coherent explanatory text. In summary, our main contributions are as follows:

\begin{itemize}
\item We introduce a novel task FNVE, which aims to generate natural language explanations for a given video news to account for the potential falsity of news videos. To the best of our knowledge, this is the first attempt to account for the falsity of multimodal news videos. 
\item We developed FakeVE, a new dataset of 2672 multimodal news video posts from FNVE that accurately explains four categories of high-prevalence fake news video aspects.
\item  We propose a Multimodal Relation Graph Transformer (MRGT) model to benchmark FakeVE, which will serve as a powerful baseline. The empirical results show that it is superior to various existing models adopted for this task.
\item We performed extensive evaluations to validate the quality of FakeVE annotations and the validity of the baseline model-generated explanations. As a byproduct, we release our code and dataset to facilitate this community.

\end{itemize}
\section{Related Work}
\subsection{Datasets} 
In recent years, research on the detection of fake news in the form of video has made remarkable progress, and the construction of several large-scale data sets has laid the foundation for this field. \citet{17} proposed that FakeSV is the largest Chinese fake news short video dataset at present, which is characterized by both news content features and social context features. \citet{16} developed the COVID-19 VTS benchmark to realize the automatic generation of misleading news videos through event manipulation and adversarial matching technology. In order to enhance data diversity, \citet{19} built an English data set FakeTT based on TikTok platform, which is closer to the real scene. In addition, \citet{24} built FMNV, a news video released by media organizations, which is the largest English news video dataset at present. However, most of the existing data set work regards fake news videos as a simple binary classification task and lacks explainable analysis of the reasons for determination, which limits users' understanding of the nature of false content.

\subsection{Generation Techniques}
At present, the multimodal content generation technology has made remarkable progress \cite{18,36,28,20}. In terms of image description generation, \citet{36} improved the description accuracy by enhancing the perception of scene changes, and \citet{28} improved the multi-modal alignment effect by using the shared encoder architecture. In the field of video generation, \citet{20} significantly improved the quality of abstract by mining cross-level feature associations. However, these technologies still have significant limitations in the field of FNVE: On the one hand, there is a lack of generative model architecture designed specifically for FNVE tasks; On the other hand, it is difficult for existing methods to distinguish and generate fine-grained false information \cite{21,22}.

 Recently, in a news video task, \citet{23} proposed to explain rumors from the perspective of modal tampering, and designed an interpretability mechanism by tracing back the decision-making process of the model. However, although these studies have made progress in using NLE to explain the model output aspect, they have mainly focused on using NLE to prove the model output of other aspects. In our task, the NLE itself is the output of a model designed to explain the falsity of news video content predictions.

\section{Proposed Dataset}
\subsection{Dataset Construction}
\noindent \textbf{Sources} This section details our efforts in developing the FakeVE dataset. As far as we know, there is still a lack of publicly available and interpreted news video datasets. Since FNVE requires fake news video posts, we explored two existing publicly available FMVD datasets, FMNV \cite{24} and FakeTT \cite{19}. The FMNV dataset, derived from the video platforms of twitter and utuber, has 2,393 news video posts containing 102 news events, including 1,500 fake news videos. The FakeTT dataset, on the other hand, consists of English news videos on the TikTok platform, with 1,991 news video posts containing 286 news events, including 1,172 fake news videos. In total, we collected a total of 2,672 fake news video posts from FMNV and FakeTT. Next, we use the instructions written by the first author below to generate an explanation for each video news post.

\noindent{\bf Aspect Inclusion:} Judging the false aspect of news video:
\begin{itemize}
	\item Contextual Dishonesty (CD) : News titles are semantic or out of context to video or audio content.
	\item Splice Tampering (ST) : Pering a news video is deliberately selected or randomly spliced, resulting in a lack of logical coherence or data unrelated to the title or audio theme.
	\item Synthetic Voiceover (SV) : news audio is synthetic speech generated by clips whose content is inconsistent with the subject matter of the video picture or the title.
	\item Contrived Absurdity (CA) : News video, audio, and titles are consistent in their multimodal content, but the overall message defies basic common sense or logic.
\end{itemize}

\noindent {\bf Annotation Scheme:} Annotators are given the following instructions to generate explanations:
\begin{itemize}
	\item Ensuring that these elements are used to explain falsity and generate an appropriate explanation according to the frame content, title, and transcript text in the video.
	\item  If a fake news video post can be explained in multiple ways, prefer a more concise, straightforward explanation.
         \item Using the contradiction to describe the  explanation and the form can be ``..., but ...'', ``..., while ...'', ``..., with on mention of ...'', and etc.
	\item The introduction of any topics that are not related to the false explanation of the news video should be avoided to ensure the pertinence and accuracy of the explanation.    
\end{itemize}

\input{tables/1}

We hired four professional annotators in different fields, who hold at least a bachelor's degree, to follow the instructions written by the first author above to ensure even quality across annotations. Each time an annotator annotates a fake news video post, they need to watch the news video and check the title. Before the formal annotation, all annotators participated in the pilot annotation test, and Cohen's Kappa coefficient \cite{25} of different annotators was 0.865, indicating good consistency and accuracy of annotations. We pay an average hourly rate to all annotators, and each annotator completes the assigned task in about 30 hours on average. Each fake news video post was rigorously vetted by two independent annotators, and for two annotators to interpret different instances of the aspect, the first author scrutinized the news video to determine the final explanation aspect. Summary statistics for the dataset are shown in Table \ref{tab:table1}.

\subsection{Dataset Analysis}

\begin{figure}[t]
	\centering
	\includegraphics[width=1\linewidth]{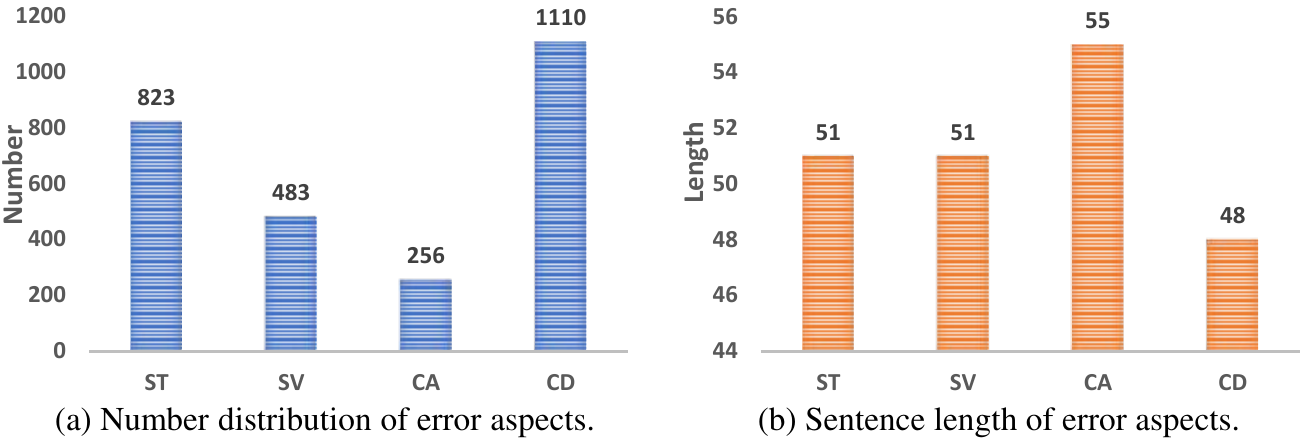}
	\caption{Statistics of four error aspects of FakeVE about length and quantity.}
	\label{fig:f2}
\end{figure}

Figure \ref{fig:f2} shows the number and sentence length distributions of four types of errors (CD, ST, SV, CA) in news production. Figure \ref{fig:f2} (a) reflects: 1) The largest number of CDS, which indicates that semantic contradictions or logical disconnects between content are the most common problem. This can be due to poor integration of information in the production process or exaggerated titles to attract attention. 2) SV is less in quantity than CD and ST. That may be because synthetic speech technology, while advancing, is still limited in its use in journalism. 3) The minimum number of cas indicates that there are relatively few cases in which the overall information violates basic common sense or logic, probably because such errors are too obvious and easy to detect and correct. Figure \ref{fig:f2} (b) reflects: 1) CD is the longest of these types of errors, which means that when there is a semantic contradiction or logical disconnect, more words are often needed to describe and explain the inconsistency. 2) CA is slightly shorter than other types, because the overall information that violates basic common sense or logical errors is more intuitive and can be understood without too much text description.

\begin{figure}[t]
	\centering
	\includegraphics[width=1\linewidth]{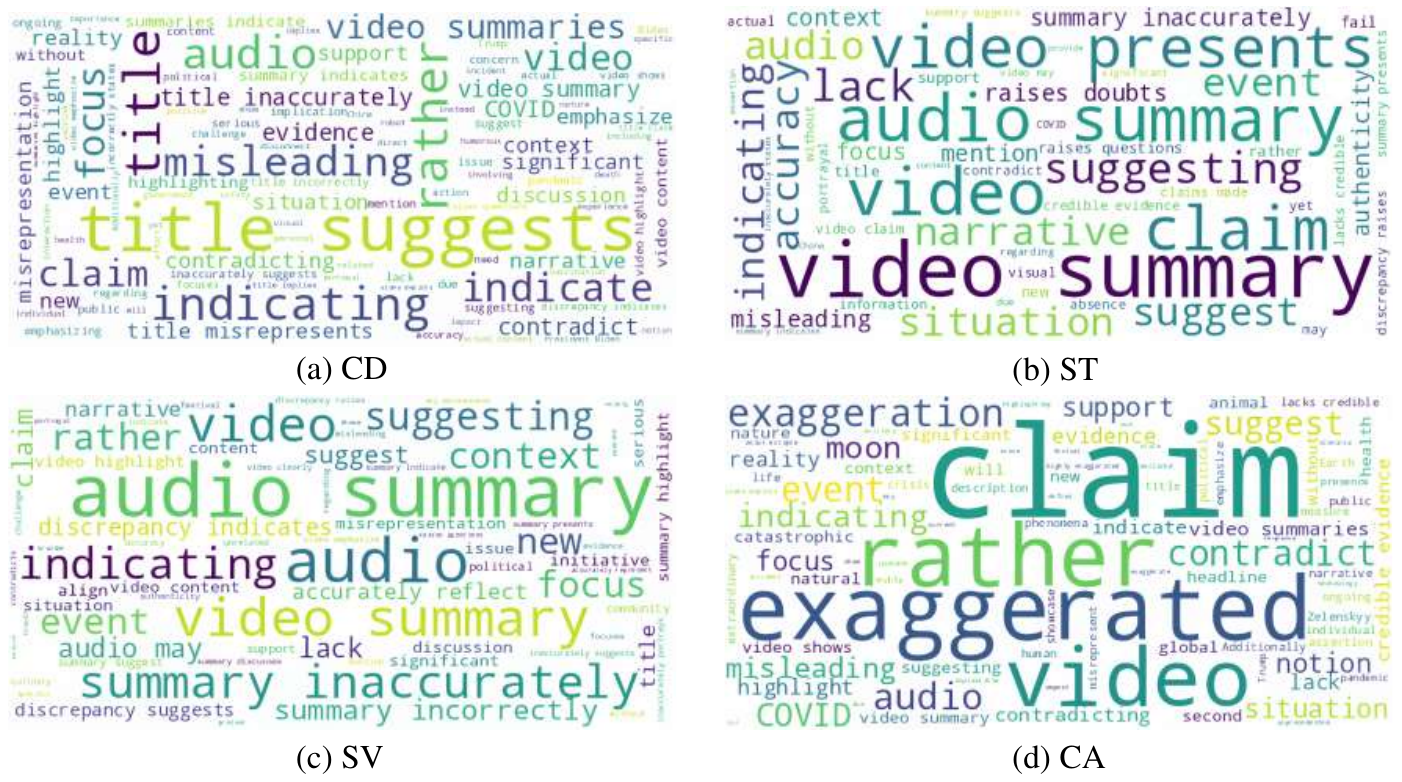}
	\caption{Word cloud of four error aspects of the explanation on FakeVE.}
	\label{fig:f3}
\end{figure}


Figure \ref{fig:f3} visually shows the specific performance of four types of errors in news (CD, ST, SV, CA) and related keywords in the form of word cloud. CD errors mainly involve inconsistent and misleading title, video and audio content, as shown by keywords such as "title", "video" and "audio", reminding news producers to ensure consistency and logical coherence between the various elements. ST error stems from improper video stitching, resulting in inaccurate and misleading information, emphasizing that video editing should be more rigorous to ensure that the screen content is consistent. SV errors involve improper use of synthetic speech and suggest caution in the application of synthetic speech technology. Although there are few CA errors, news content contrary to common sense or logic still need to be vigilant, such as "inaccurately" and other keywords, news production team should strengthen the audit of the content, in line with common sense and logic.

\subsection{Evaluations on Datasets}

\subsubsection{Evaluation Metrics}
In this section, we assess the explanatory quality of the proposed FakeVE. We employ G-Eval \cite{27}, a referenceless evaluation method based on LLM that excels at evaluating text quality from multiple perspectives, even without reference text. Therefore, we use explanatory quality \cite{15} based on four metrics widely used in human assessment: \textit{Persuasiveness}, \textit{Informativeness}, \textit{Soundness}, and \textit{Readability}. A 5-point Likert scale is used, where 1 is the worst and 5 is the best. Indicators are defined as:

\begin{itemize}
	\item \textit{Persuasiveness} (P) evaluates whether the model's explanation is convincing, with ratings ranging from 1 (not persuasive) to 5 (very persuasive).
	\item \textit{Informativeness} (I) evaluates whether the explanation provides new information, such as explanatory background and additional context, on a scale of 1 (not informative) to 5 (very informative).
		\item \textit{Soundness} (S) describes the validity and logic of the explanation on a scale of 1 (not sound) to 5 (very sound).
		\item \textit{Readability} (R) assesses whether the explanation follows appropriate grammatical and structural rules, and whether the sentences in the explanation are appropriate and easy to follow, on a scale of 1 (poor) to 5 (excellent).
\end{itemize}

\subsubsection{Evaluation Baselines}
In recent years, Multimodal Large Language Models (MLLMs) \cite{3,26} have demonstrated remarkable reasoning ability in natural language understanding and generation tasks, especially in complex tasks. Therefore, we chose MLLMs as a baseline comparison: (1) Qwen2-VL \cite{34} is an advanced MLLM capable of simultaneously processing visual and linguistic information to achieve joint understanding and generation of images and text. (2) GPT-4o \cite{29}, as the most advanced LLM currently available, demonstrates excellent capabilities in multi-modal understanding and complex reasoning tasks.

We use MLLMs causal reasoning capabilities to interpret fake news videos. Given a news title $T$ and a video frame $F$, the audio is converted to transcript as $A$ and labeled $Y$. We have planned a prompt template $P$. We use it to prompt MLLMs to generate explanation $E$, which elicits reasons for how $Y=False$ can be inferred from the interaction of multimodal information from news videos. We design $P$ as:

\par
\textit{Given a news video post containing a news title \textlangle T\textrangle, a video frame \textlangle F\textrangle, and an audio transcript \textlangle A\textrangle. If the veracity of the current news video post is \textlangle Y\textrangle, please analyze the reason for the falsehood from the point of view of the logical contradiction of the video content and the consistency of the news title with the video, and give a short and effective explanation.}

\begin{figure}[t]
	\centering
	\includegraphics[width=1\linewidth]{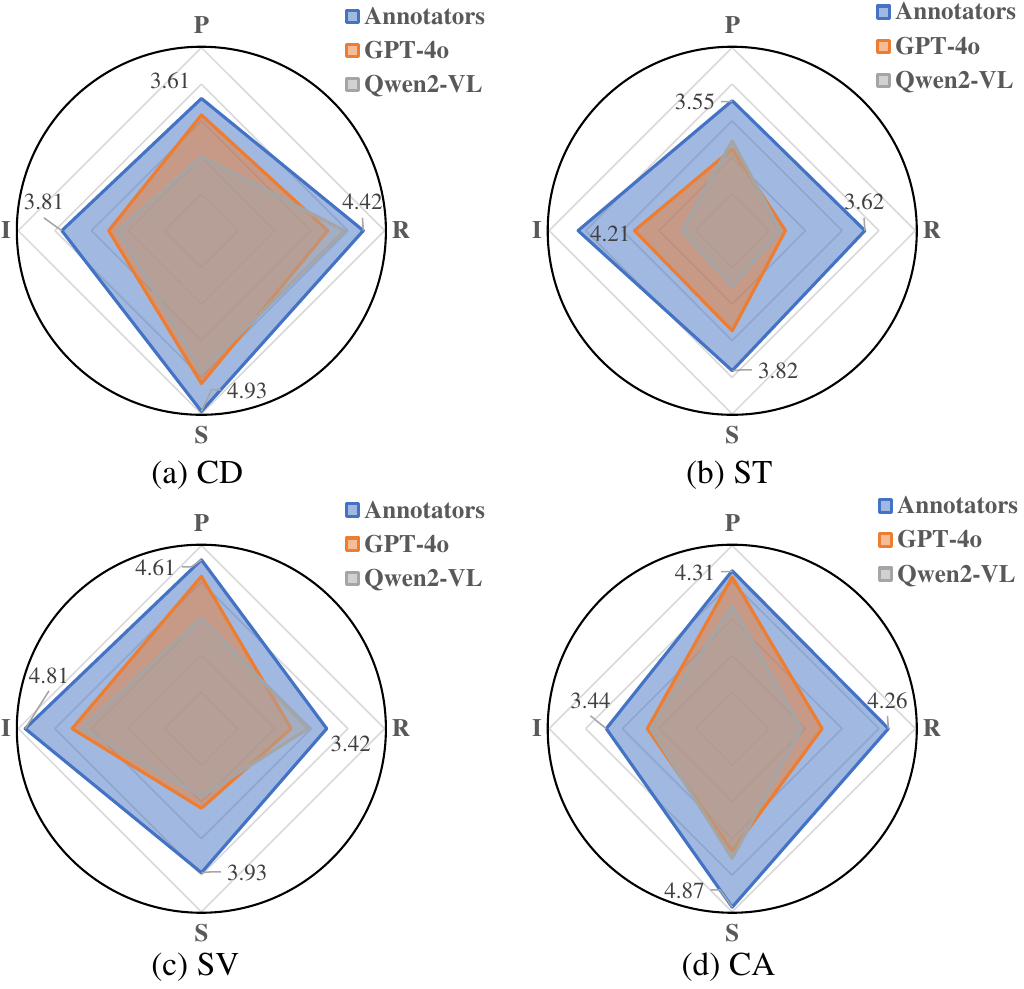}
	\caption{Evaluation results of explanation quality using a 5-Point Likert scale rating on four false aspects of the FakeVE.}
	\label{fig:f4}
\end{figure}

\begin{figure*}[h]
	\centering
	\includegraphics[width=1\linewidth]{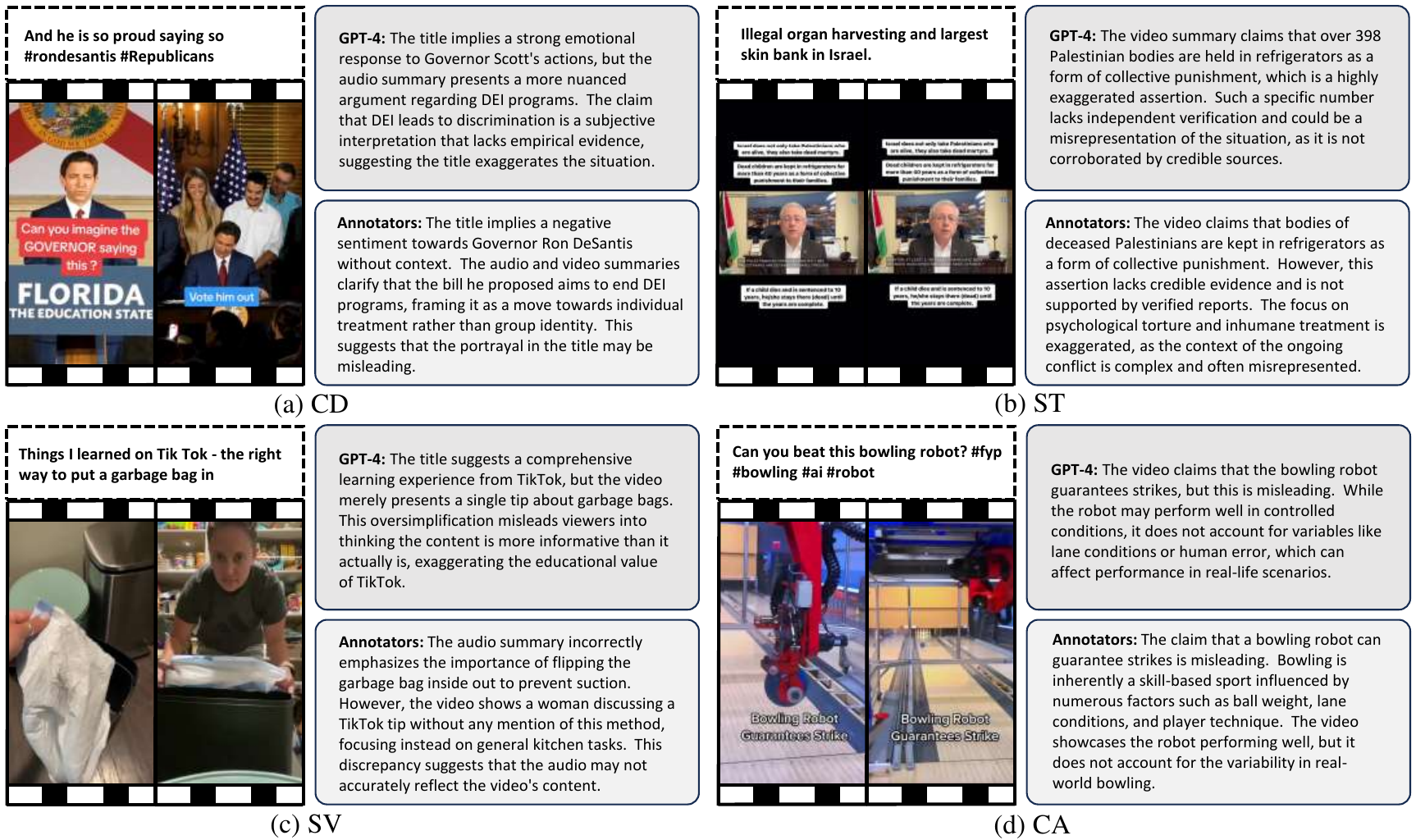}
	\caption{Examples of the explainable aspects of four different errors analyzed by GPT-4 and the human annotators, showing similarities and differences in explanation approaches, and their insights into the nature of these errors.}
	\label{fig:f5}
\end{figure*}

\subsubsection{Evaluation Analysis}
Figure \ref{fig:f4} shows the quality of explanation generated by manual annotations evaluated by G-Eval versus MLLMs (GPT-4 and Qwen2-VL). The results show that: 1) In terms of CD, manual annotation performs well in the four indicators of persuasion, information, logic and readability, indicating the high quality of its explanation. GPT-4 and Qwen2-VL have some performance, but there is still a significant gap between them and manual annotation. 2) In terms of ST, manual annotation also performs well, showing stylistic consistency and high quality of explanation. GPT-4 score is slightly higher than Qwen2-VL, indicating that it is slightly better in style consistency. 3) In terms of SV and CA, manual annotations maintain high scores, indicating a high degree of accuracy and reliability in their explanation. In contrast, GPT-4 and Qwen2-VL scored between 3 and 4 on the same assessment criteria, which means that there is still a lot to be done in terms of information and logic.

Examples of four error aspects of annotators and GPT-4 are shown in Figure \ref{fig:f5}, respectively. We found that annotators were more accurate and nuanced in understanding and interpreting the video content than GPT-4. Specifically, in the CD aspect, the annotators more accurately pointed out the misleading title and explained the actual intent of the audio and video content; In the ST aspect, annotators more accurately pointed out the lack of evidence and exaggerated focus of the video content, and explained the complexity of the background; In the SV aspect, annotators more specifically point out the inconsistency between the audio summary and the video content, and explain the actual content of the video; On the CA side, annotators explain the intricacies of bowling in more detail and point out the differences between video presentations and real-world applications. From these analyses, we can see that annotators not only cover the false explanations of GPT-4, but also provide more detailed and accurate explanations and background information.

\section{Proposed Model}
\begin{figure*}[t]	
	\centering
	\includegraphics[width=1\linewidth]{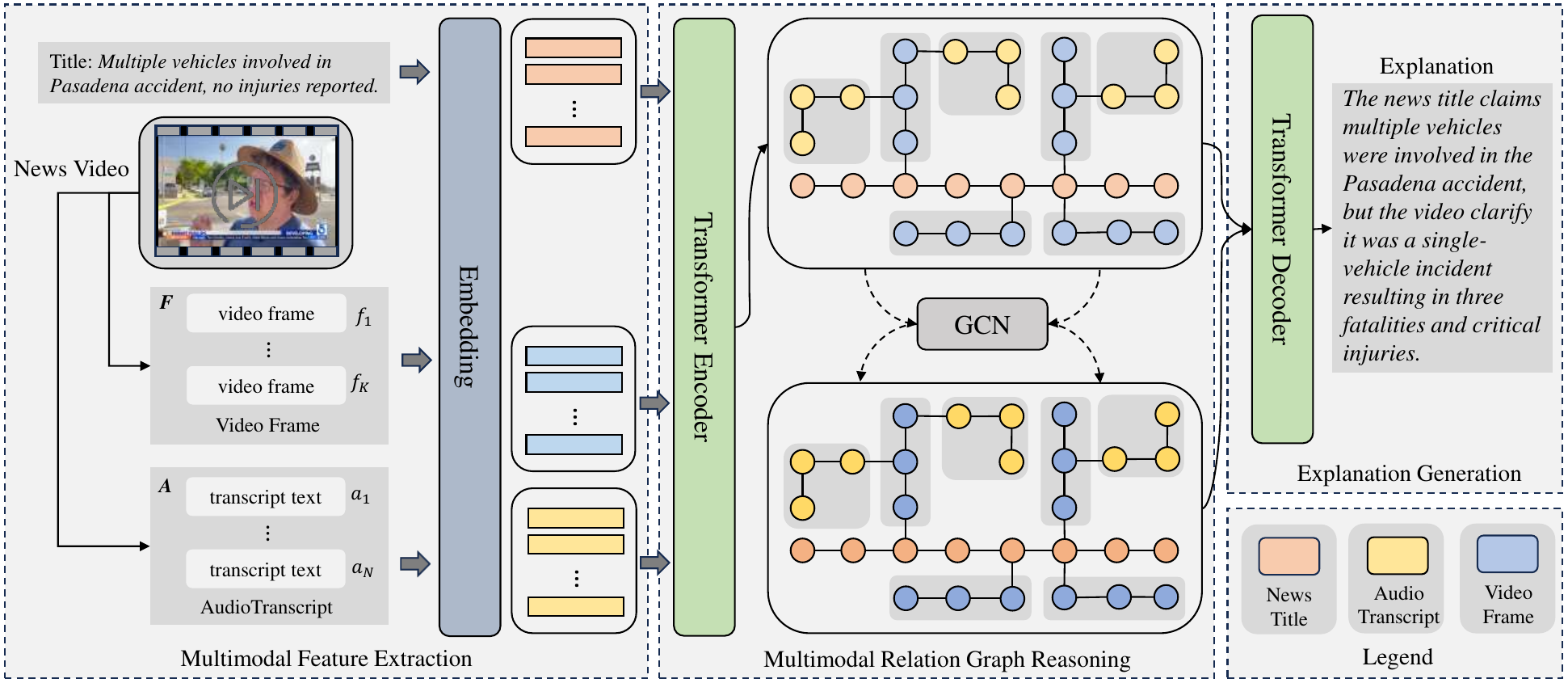}
	\caption{The architecture of the proposed MRGT, which consists of three key components: Multimodal Feature Extraction, Multimodal Relation Graph Reasoning and Explanation Generation.}
	\label{fig:f6}
\end{figure*}

To benchmark FakeVE, we propose MRGT, a multimodal Transformer based Encoder-Decoder framework, as shown in Figure \ref{fig:f6}. We formally define FNVE as follows: For a given fake news video post $V = \{T, F, A\}$, where $F$ represents the video frame, $T$ is the news title and $A$ is the audio transcript. Our goal is to reveal the fakery of news videos by generating a natural language explanation $E[{e_1},{e_2}, \cdots {e_N}]$, where $E$ represents the target explanation text consisting of $N$ tokens.

\subsection{Multimodal Feature Extraction}
The news titles are usually the audience's first impression of the news content. An accurate and objective title can guide the audience in forming reasonable expectations about the news content \cite{30}. Therefore, we first use the news title $T\left[ {{t_1},{t_2}, \cdots ,{t_L}} \right]$, where ${t_i}$ is the $i$-th token in the news title. We observe that key objects in video frames often contain important semantic information, which is crucial for identifying fake content. For the visual mode, we extract the video frame $F\left[ {{f_{\rm{1}}},{f_{\rm{2}}}, \cdots ,{f_K}} \right]$ from the uniform sample of each video, where ${f_i}$ is the $i$-th video frame in the video. In fake news videos, audio can convey the emotional semantics and live reports of the video's creators, and these verbal information are critical to judging the veracity and intent of the video content. For audio modes, we convert the audio into transcript denoted as $A\left[ {{a_{\rm{1}}},{a_{\rm{2}}}, \cdots ,{a_N}} \right]$, where ${a_i}$ is the $i$-th frame audio transcript in the video.

\subsection{Multimodal Relation Graph Reasoning}
So far, we obtained three kinds of modal  sources for news videos: the news title $T$, the audio transcript $A$, and the video frame $F$. To extract the embedded features, we turn to the Transformer \cite{32} encoder, which has shown convincing success on various natural language processing tasks. We first concatenate them into a sequence of tokens, denoted as $X$, and then feed $X$ into the Transformer encoder $TE$ as follows:
\begin{equation}
	H = TE(X),
\end{equation}
where $H \in \mathbb{R}{^{S \times D}}$ is the coded representation matrix, each column corresponds to a token, and $S$ is the total number of tokens in $X$.

In fact, there are rich relations among the three modal sources available for false inference and corresponding explanation generation. For example, the modal correlation between the input news title and audio transcript tags can facilitate cross-modal inconsistency uncovering; The modal correspondence between the input video frame and audio transcript data can help to mine modal intra-modal inconsistencies. Hence, for each sample $V$, we construct a multimodal relation graph $G$ to comprehensively capture the above modal relations. Corresponding to $S$ tokens in $X$, it can be divided into three categories: news title nodes, audio transcript nodes, and video frame nodes.

As shown in Figure \ref{fig:f6}, the edges of this graph are defined according to the modal relations between these nodes. We establish the following rules to construct the edges: 1) We link semantically related text nodes by adding an edge in the news title, audio transcript and video frame by introducing the dependency tree relations. 2) To capture the modal relations of cross-modal, we establish an edge between all video frames and their news title tags with the highest modal  similarity measured by cosine similarity. 3) To capture the modal relations within the video modality, we link the visual nodes by adding an edge between the video frame and audio transcript. Formally, let $A \in \mathbb{R}{^{S \times S}}$ denote the adjacency matrix of the multimodal relational graph we constructed.

Thereafter, we resort to the commonly used GCN for fake news video inference. Specifically, suppose we adopt GCN $L$ layer, all node representations are updated iteratively as follows:
\begin{equation}
{{\bf{G}}_l} = {\mathop{\rm Re}\nolimits} LU\left( {\widetilde {\bf{A}}{{\bf{G}}_{l - 1}}{{\bf{W}}_l}} \right),l \in [1,L],
\end{equation}
where $\widetilde {\bf{A}} = {({\bf{D}})^{ - \frac{1}{2}}}{\bf{A}}{({\bf{D}})^{ - \frac{1}{2}}}$ is the normalized symmetric adjacency matrix and $D$ is the degree matrix of $A$. ${{\bf{G}}_l}$ is the representation of the node obtained in layer $l$ GCN, where ${{\bf{G}}_0} = {\bf{H}}$ is the initial node representation.

The final node obtained by the GCN indicates that the ${G_L}$ should absorb rich modal  information from its related nodes and be used as input for the following falsity explanation generation. We also introduce a residual connection \cite{33} to generate spurious explanations. Specifically, we first fuse the initial, final node representation as follows:
\begin{equation}
Z = {\bf{H}} + {{\bf{G}}_L},
\end{equation}
where $Z \in \mathbb{R}{^{S \times D}}$ denotes the fusion node representation.

\subsection{Explanation Generation}

In explanation generation, we feed $Z$ to the decoder $TD$ of the pre-trained Transformer. The decoder works in an autoregressive fashion, that is, it generates the next words by considering all previously decoded outputs as follows:

\begin{equation}
{\widehat {\bf{y}}_t} = TD\left( {Z,{{{\rm{\hat Y}}}_{ < {\rm{t}}}}} \right),
\end{equation}
where $Z \in \mathbb{R}{^{S \times D}}$, $t \in \left[ {1,{N_y}} \right]$ and ${\widehat {\bf{y}}_t} \in\mathbb{R} {^{|{\cal V}|}}$ are the $t$-th token probability distributions of spurious explanation. ${\hat Y_{ < t}}$ refers to the previously predicted $t$-1 labeling.

To optimize the MRGT generation, we again apply the standard cross entropy loss function as follows:
\begin{equation}
{{\cal L}_{Gen}} =  - 1/{N_y}\sum\limits_{i = 1}^{{N_y}} {\log } \left( {{{\widehat {\bf{y}}}_i}[t]} \right)
\end{equation}
where ${\widehat {\bf{y}}_i}[t]$ is the element of ${\widehat {\bf{y}}_i}$ corresponding to the $i$-th token of the generated explanation, and ${N_y}$ is the total number of tokens in the generated truth explanation $Y$.

\input{tables/2}
\section{Experiments}
\subsection{Comparative Systems}

We chose Multimodal Large Language Models (MLLMs) and small pre-training models (SPMs) for comparison:

MLLMs: 1) Qwen2-VL \cite{34}: A series of multimodal large language models developed by Qwen team of Alibaba Cloud, with advanced image and video understanding capabilities. 2) LLaVA \cite{35}: is a large-scale multimodal model of end-to-end train designed to understand and generate content based on visual input images and text instructions. 3) GPT-4o \cite{29}: The latest version of GPT-4 Omni, an LLM developed by OpenAI for quick invocation.

SPMs: 1) NCT \cite{36}: A neighborhood contrast converter is proposed to improve the model's ability to perceive various changes in complex scenes and to recognize complex syntactic structures. 2) HAAV \cite{28}: each view in the multimode is encoded independently by a shared encoder, and the contrast loss is incorporated in the encoded view in a novel way. 3) AMFM \cite{20}: dynamic enhancement of pre-trained visual features by learning the underlying visual relationship between frame level and video level embedment.

\subsection{Experimental Setup}

We conduct experiments on FakeVE and use an 8:1:1 split to create training (2138), validation (267), and test (267) sets. We selected several standard metrics for evaluating the performance of generated explanations. These include 1) BLEU \cite{37} (BLEU-1, BLEU-2, BLEU-3, BLEU-4), which measures the similarity between the generated text and the reference text, 2) ROUGE \cite{38} (ROUGE-1, ROUGE-2, ROUGE-L), which evaluates the similarity between the generated summary and the reference summary, 3) METEOR \cite{31}, which assesses explanation quality using synonym matching strategies, and 4) SentBERT\cite{26}, which uses Sentence-BERT to embed modal  similarity between reference and generated explanations in the space.

We adopt the bart-base model \cite{32} provided by huggingface as the backbone of our model. In practice, we concatenate all modal  texts, unified to 312 by padding or truncation operations. The feature dimension D is set to 768 and the maximum number of intercepts we allow to be extracted from the image is set to 55. We used AdamW \cite{3} as the optimizer and set the learning rate to 1e-3 for the GCN layer and 1e-4 for BART. The batch size is set to 16 and the maximum number of epochs for model training is set to 15.

\input{tables/3}

\subsection{Experimental Results}
Table \ref{tab:table2} shows the results of benchmarking the various baseline models on the FakeVE generation task, while comparing the MLLM performance of the zero sample. We observe that: 1) GPT-4o has higher scores than Qwen2-VL and LLaVA, and even exceeds the fine-tuning model PTSN, which indicates that GPT-4o has stronger language generation ability and stronger generalization ability on FNVE tasks. 2) The scores of PTSN and HAAV are lower than those of AMFM and MRGT because they focus on multimodal representation of vision and language and lack sufficient language generation capability and capturing nuance of text in plain text generation tasks. 3) AMFM is competitive in video summary generation and performs better on M and Sent-B scores, indicating that the generated explanation is highly consistent with the video content at the semantic level. 4) MRGT has the highest score in BLEU and Rouge indicators compared with other models, which indicates that the explanatory sentences generated by MRGT can not only capture video content more accurately, but also generate explanations with higher overall similarity and coherence.

\begin{figure}[t]
	\centering
	\includegraphics[width=1\linewidth]{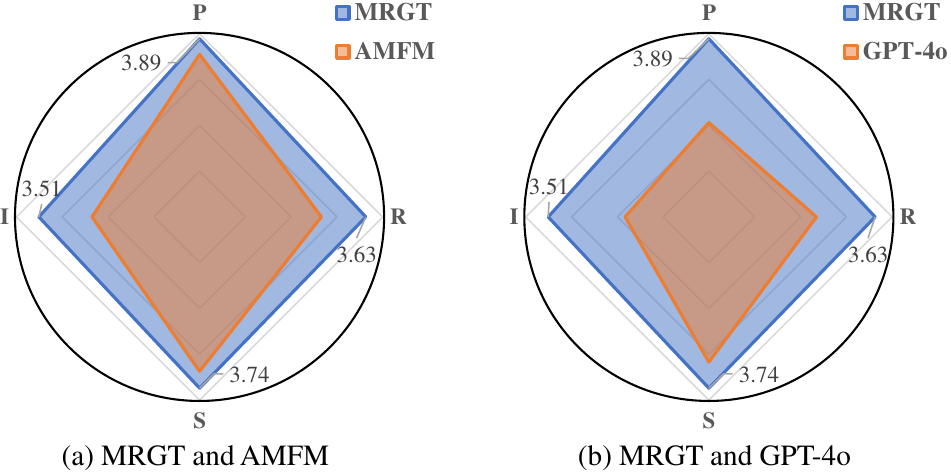}
	\caption{Evaluation results of explanation quality using a 5-Point Likert scale rating on FakeVE.}
	\label{fig:f7}
\end{figure}

\begin{figure*}[h]
	\centering
	\includegraphics[width=1\linewidth]{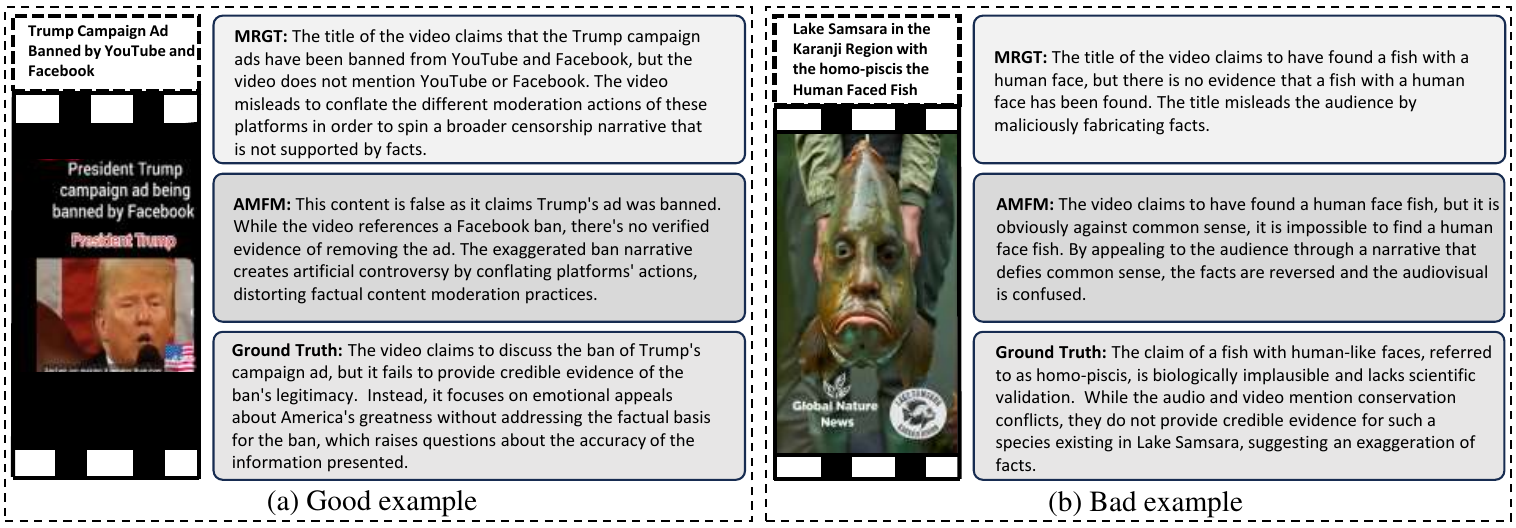}
	\caption{Examples of good and bad explanations produced by MRGT and AMFM in the FakeVE dataset.}
	\label{fig:f8}
\end{figure*}

\subsection{Ablation Study}
We designed several model variants to evaluate the role of different information sources in FNVE generation. First, we remove the video frame information (w/o Frame) to understand the contribution of the video frame in the model. Second, we also excluded audio transcript data (w/o Transcript) to analyze the role of audio transcript in explanation generation. Finally, to verify the necessity of multi-source graphs in spurned inference, we remove the Graph construction part (w/o Graph). As shown in Table \ref{tab:table3}, model variants that remove the audio transcript (w/o Transcript) and video frame (w/o Frame) show a significant decline in performance compared to MRGT. This is because the video frame and audio transcript are the core information sources for spurious reasoning, and their absence naturally results in performance degradation. More importantly, MRGT outperforms variants that remove the w/o Graph, which highlights the graph structure's indispensability in capturing the falsity of multimodal news video posts.

\subsection{Evaluation on Explanation}
Figure \ref{fig:f7} shows the generation performance of the three models MRGT, AMFM and GPT-4o in the four key indicators of persuasion (P), information (I), readability (R) and logic (S). We can see that: 1) Because the MRGT model absorbs rich semantic information in relevant nodes by using Graph, it has achieved excellent results in the four indicators. This feature makes MRGT generation explanation show extraordinary strength in constructing coherent narrative and transmitting rich information generation. 2) The AMFM model, while excellent in terms of information, is slightly inferior in terms of persuasion and readability. The resulting text may fall short in capturing the reader's attention and maintaining reading flow. 3) The GPT-4o model, on the other hand, shows outstanding ability in persuasion, and the text generated by it is often very persuasive. However, in terms of information and readability, GPT-4o is relatively weak. This may be related to the fact that the model focuses more on visual language processing, which may lead to the excessive pursuit of visual effects in text generation.

\subsection{Case Study}
In Figure \ref{fig:f8}, we show two examples of FNVE, each including MRGT, AMFM explanation, and Ground Truth. In example (a), the case for generating an explanation good is shown. MRGT's explanation is more comprehensive and accurate, pointing out not only the misleading claim by the video's title that Trump's ads were banned on YouTube and Facebook, but also how the video's content misleads viewers and creates broader censorship rhetoric. In contrast, AMFM, while also pointing out errors in the content, mainly focuses on the false claim that the AD was banned and does not delve into the inconsistency between the title and the content. In example (b), the case of generating explanation bad is shown. Neither MRGT's nor AMFM's explanations fully pinpoint the heart of the problem. MRGT simply points out the misleading claim of the title that a fish with a human face has been found in Lake Laksamara, while AMFM, while pointing out the common sense impossibility, does not delve into the misleading nature of the title and content.

\section{Conclusion and Potential Applications}
In this paper, we propose a novel fake news video explanation task, which aims to analyze falsehood in news videos by combining titles and video content. To do this, we built a brand new dataset, FakeVE, with 2,672 fake news video posts annotated with reference explanations in natural language (English) sentences. To verify its validity, we designed a robust baseline model, MRGT, which introduces multimodal diagrams for benchmarking FakeVE datasets. Through a series of evaluations, it was found that MRGT demonstrated better performance than other baseline models.

In addition to the work in this paper, the explanations provided by FakeVE have significant spillover value: 1) Fine-grained evidence chains. The false aspects clearly labeled in the explanation can be directly used to train the model to recognize complex tampering techniques. 2) Public awareness is improved. We can combine explanation to transform professional analysis into popular science materials that the public can understand, and effectively reduce the threshold of fake news identification by revealing manipulation techniques such as clickbait and context deception. 3) Model security improvement. The potential of adversarial testing can expose the weakness of multimodal generation models and point out the direction of model robustness optimization.


\bibliography{custom}

\appendix

\end{document}

%% file: tables/1.tex
\begin{table}[h]
	\centering
    	\caption{Statistics of the FakeVE dataset, where ``Avg.''. ``Dur.'' and ``Exp.'' refer to ``Average'', ``Duration'' and ``Explanation'', respectively.}
        
	\resizebox{\columnwidth}{!}{
		\begin{tabular}{c c c c c}
        \hline
			{\bf Split} & {\bf \#of News} & {\bf Avg.Title} & {\bf Avg. Dur (s)}&{\bf Avg. Exp} \\
            \hline
			Train&	2138 & 21.40 & 61.23 & 49.76\\
			Val&	267 & 16.17 & 63.45 & 50.50\\
			Test&267 & 15.37 & 60.32 & 50.08\\
			\hline
			Total & 2672 & 20.27 & 61.78 & 49.86\\
			\hline
		\end{tabular}
	}
	\label{tab:table1}
\end{table}

%% file: tables/2.tex
\begin{table*}[htbp]
	\centering
            	\caption{Results of comparison among different models on FakeVE dataset, where the best results are in bold. B@1, B@2, B@3, B@4, M, R-1, R-2, R-L and Sent-B are short for BLEU-1, BLEU-2, BLEU-3, BLEU-4, METEOR, ROUGE-1, ROUGE-2, ROUGE-L and SentBERT.}
     \setlength{\tabcolsep}{9pt}{
		\begin{tabular}{lccccccccc}
			
			\hline
		\multicolumn{1}{c}{\multirow{2}{*}{\textbf{Method}}} 
			& \multicolumn{4}{c }{\textbf{BLEU}} & \multicolumn{1}{c}{\multirow{2}{*}{\textbf{M}}} & \multicolumn{3}{c}{\textbf{Rouge}}&\multicolumn{1}{c}{\multirow{2}{*}{\textbf{Sent-B}}}  \\
			\cmidrule(lr){2-5} \cmidrule(lr){7-9}
  & B@1&	B@2&	B@3&	B@4&	&	R-1&	R-2&	R-L& \\
	\cmidrule(lr){1-10}
	\multicolumn{10}{c}{\textit{MLLM-based approach}}\\
	\cmidrule(lr){1-10}
Qwen2-VL & 25.94 & 11.92 & 7.76 & 5.66 & 75.89 & 28.84 & 8.66 & 12.94 & 69.23\\
LLaVA & 29.32 & 15.42 & 9.13 & 6.21 & 79.75 & 31.34 & 9.27 & 16.60 & 74.02\\
GPT-4o & 32.59 & 17.33 & 11.23 & 6.70 & 82.84 & 32.57 & 9.78 & 17.8 & 77.79\\
	\cmidrule(lr){1-10}
\multicolumn{10}{c}{\textit{Fine-tuning approach}}\\
\cmidrule(lr){1-10}
	NCT & 33.12 & 16.78 & 8.45 & 4.12 & 83.34 & 27.89 & 7.12 & 18.78 & 72.12 \\
	HAAV & 34.45 & 16.90 & 9.23 & 4.78 & 84.67 & 28.12 & 8.90 & 18.45 & 75.67 \\
	AMFM & 35.67 & 18.89 & 10.01 & 6.12 & 86.68 & 33.45 & 10.89 & 22.67 &78.12 \\
	MRGT & \textbf{39.39} & \textbf{22.44} & \textbf{13.08} & \textbf{8.21} & \textbf{91.62} & \textbf{38.07} & \textbf{12.34} & \textbf{27.21} & \textbf{84.66} \\

	 \hline			
		
		\end{tabular}%
        }
		\label{tab:table2}%
	    \vspace{-0.3cm}
\end{table*}%

%% file: tables/3.tex
\begin{table}[h]
	\centering
  \caption{Ablation study on different modalities on the FakeVE dataset.}
\setlength{\tabcolsep}{3pt}{
		\begin{tabular}{c c c c c c }
			\hline

		Model &	B@1&	B@4	&M&	R-L &Sent-B\\
	\hline
        w/o Frame & 34.36 & 5.43 & 83.44 & 23.16 & 75.90\\
		w/o Transcript & 34.62 & 6.65 & 83.75 & 23.35  & 76.23\\
        w/o Graph & 35.43 &6.32 & 84.52 & 22.60 & 78.75\\
		MRGT & \textbf{39.39} & \textbf{8.21} & \textbf{91.62} & \textbf{27.21}& \textbf{84.66}\\
			\hline
		\end{tabular}
	}
	\label{tab:table3}
\end{table}